\documentclass[11pt,a4paper]{article}

\usepackage[hyperref]{naaclhlt2019}
\usepackage{graphicx}
\usepackage{xcolor}
\definecolor{color1}{rgb}{0.4, 0., 0.} 
\definecolor{color2}{rgb}{0.6, 0.4, 1} 

\usepackage{times}
\usepackage{latexsym}
\usepackage[utf8]{inputenc}
\usepackage{xcolor}
\usepackage{linguex}
\usepackage{rotating}
\usepackage{float}
\usepackage{booktabs}
\usepackage{multirow} 
\usepackage{url}
\usepackage{hyperref}
\usepackage{amsmath}
\usepackage{fancyhdr}
\usepackage{fancyheadings}

\usepackage[margin=1in,headsep=.2in]{geometry}

\def\aclpaperid{1562X} 

\aclfinalcopy

\title{Studying the Inductive Biases of RNNs\\ with Synthetic Variations of Natural Languages}

\author{Shauli Ravfogel\textsuperscript{1} \;\;\; Yoav Goldberg\textsuperscript{1,2} \;\;\; Tal Linzen\textsuperscript{3} \\
\textsuperscript{1}Computer Science Department, Bar Ilan University \\
\textsuperscript{2}Allen Institute for Artificial Intelligence \\
\textsuperscript{3}Department of Cognitive Science, Johns Hopkins University \\
  {\tt \{shauli.ravfogel, yoav.goldberg\}@gmail.com, tal.linzen@jhu.edu}
  }

\newcommand{\yg}[1]{\textcolor{red}{[YG: #1]}}
\newcommand{\sr}[1]{\textcolor{blue}{SR: #1}}
\newcommand{\tl}[1]{\textcolor{brown}{[TL: #1]}}

\renewcommand{\yg}[1]{}
\renewcommand{\sr}[2]{}
\renewcommand{\tl}[3]{}

\date{}

\begin{document}

\maketitle

\thispagestyle{fancy}

\begin{abstract}
How do typological properties such as word order and morphological case marking affect the ability of neural sequence models to acquire the syntax of a language? Cross-linguistic comparisons of RNNs' syntactic performance (e.g., on subject-verb agreement prediction) are complicated by the fact that any two languages differ in multiple typological properties, as well as by differences in training corpus. We propose a paradigm that addresses these issues: we create synthetic versions of English, which differ from English in one or more typological parameters, and generate corpora for those languages based on a parsed English corpus. We report a series of experiments in which RNNs were trained to predict agreement features for verbs in each of those synthetic languages. Among other findings, (1) performance was higher in subject-verb-object order (as in English) than in subject-object-verb order (as in Japanese), suggesting that RNNs have a recency bias; (2) predicting agreement with both subject and object (polypersonal agreement) improves over predicting each separately, suggesting that underlying syntactic knowledge transfers across the two tasks; and (3) overt morphological case makes agreement prediction significantly easier, regardless of word order. 

\end{abstract}

\section{Introduction}
\setlength{\Exlabelwidth}{0.25em}
\setlength{\SubExleftmargin}{1.35em}
The strong performance of recurrent neural networks (RNNs) in applied natural language processing tasks has motivated an array of studies that have investigated their ability to acquire natural language syntax without syntactic annotations; these studies have identified both strengths \cite{linzen2016assessing, giulianelli2018hood,gulordava2018LMagreement,kuncoro2018lstms,vanschijndel2018gardenpath,wilcox2018fillergap} and limitations \cite{chowdhury2018rnn,marvin2018targeted,wilcox2018fillergap}.

\begin{table*}[t]
\refstepcounter{table}\label{tbl:output-sample}
  \resizebox{\textwidth}{!}{\begin{minipage}{\textwidth}

\resizebox{\textwidth}{!}{
\begin{tabular}{lll}
\\
\textbf{Original} & & {\color{red} they} \textcolor{orange}{say} {\color{red} the broker} \textcolor{orange}{took} {\color{blue} them} out for lunch frequently . \\
& & \qquad \textit{({\color{red}subject};  {\color{orange} verb}; {\color{blue}object})}         
\\\midrule                                         \textbf{Polypersonal agreement}   & & they say\textcolor{cyan}{kon} the broker took\textcolor{brown}{kar}\textcolor{violet}{ker} them out for lunch frequently .\\
& & \qquad \textit{(\textcolor{cyan}{kon}: plural subject; \textcolor{brown}{kar}: singular subject; \textcolor{violet}{ker}: plural object)}\\ \midrule
\textbf{Word order variation} &
 SVO & {\color{red} they} \textcolor{orange}{say} {\color{red} the broker} \textcolor{orange}{took} out frequently {\color{blue} them} for lunch . \\
& SOV & {\color{red} they} {\color{red} the broker} {\color{blue} them} \textcolor{orange}{took} out frequently for lunch \textcolor{orange}{say} . \\
& VOS & \textcolor{orange}{say} \textcolor{orange}{took} out frequently {\color{blue} them} {\color{red} the broker} for lunch {\color{red} they}. \\ 
& VSO & \textcolor{orange}{say} {\color{red} they} \textcolor{orange}{took} out frequently {\color{red} the broker} {\color{blue} them} for lunch . \\
& OSV & {\color{blue} them} {\color{red}the broker} \textcolor{orange}{took} out frequently for lunch {\color{red} they} \textcolor{orange}{say} . \\
 & OVS & {\color{blue} them} \textcolor{orange}{took} out frequently {\color{red} the broker} for lunch {\color{green} \textcolor{orange}{say}}  {\color{red} they}

.

\\ \midrule

\textbf{Case systems}         &
Unambiguous & they\textcolor{cyan}{kon} say\textcolor{cyan}{kon} the broker\textcolor{brown}{kar} took\textcolor{brown}{kar}\textcolor{violet}{ker} they\textcolor{violet}{ker} out for lunch frequently .\\ 
& & \qquad \textit{(\textcolor{cyan}{kon}: plural subject; \textcolor{brown}{kar}: singular subject; \textcolor{violet}{ker}: plural object)}\\

& Syncretic & they\textcolor{cyan}{kon} say\textcolor{cyan}{kon} the broker\textcolor{magenta}{kar} took\textcolor{magenta}{karkar} they\textcolor{magenta}{kar} out for lunch frequently .\\ 
& & \qquad \textit{(\textcolor{cyan}{kon}: plural subject; \textcolor{magenta}{kar}: plural object/singular subject)} \\
& Argument marking & they\textcolor{color1}{ker} say\textcolor{color1}{ker} the broker\textcolor{color2}{kin} took\textcolor{color1}{ker}\textcolor{color2}{kin} they\textcolor{color1}{ker} out for lunch frequently .\\
& & \qquad (\textit{\textcolor{color1}{ker}: plural argument; \textcolor{color2}{kin}: singular argument})
\end{tabular}
}
\caption*{Figure 1:  The sentences generated in our synthetic languages based on an original English sentence. All verbs in the experiments reported in the paper carried subject and object agreement suffixes as in the polypersonal agreement experiment; we omitted these suffixes from the word order variation examples in the table for ease of reading. }
      \end{minipage}}

\end{table*}

Most of the work so far has focused on English, a language with a specific word order and relatively poor morphology. 
Do the typological properties of a language affect the ability of RNNs to learn its syntactic regularities? Recent studies suggest that they might. \citet{gulordava2018LMagreement} evaluated language models on agreement prediction in English, Russian, Italian and Hebrew, and found worse performance on English than the other languages. In the other direction, a study on agreement prediction in Basque showed substantially \textit{worse} average-case performance than reported for English \cite{ravfogel2018basque}. 

Existing cross-linguistic comparisons are difficult to interpret, however. Models were inevitably trained on a different corpus for each language. The constructions tested can differ across languages \cite{gulordava2018LMagreement}. Perhaps most importantly, any two natural languages differ in a number of typological dimensions, such as morphological richness, word order, or explicit case  marking. This paper proposes a controlled experimental paradigm for studying the interaction of the inductive bias of a neural architecture with particular typological properties. Given a parsed corpus for a particular natural language (English, in our experiments), we generate corpora for synthetic languages that differ from the original language in one of more typological parameters \cite{chomsky1981lectures}, following \newcite{wang2016galactic}. In a synthetic version of English with a subject-object-verb order, for example, sentence \ref{original} would be transformed into \ref{sov}:

\ex.\a.\label{original} The man eats the apples.
    \b.\label{sov}The man the apples eats.

 We then train a model to predict the agreement features of the verb; in the present paper, we focus on predicting the plurality of the subject and the object (that is, whether they are singular or plural). The subject plurality prediction problem for \ref{sov}, for example, can be formulated as follows:

    \ex.\label{sov-q}The man the apples $\langle$singular/plural subject?$\rangle$.

We illustrate the potential of this approach in a series of case studies. We first experiment with polypersonal agreement, in which the verb agrees with both the subject and the object (\S\ref{polypersonal-section}). We then manipulate the order of the subject, the object and the verb (\S\ref{order-section}), and experiment with overt morphological case (\S\ref{cases-section}). For a preview of our synthetic languages, see Figure~\ref{tbl:output-sample}.

\section{Setup}

\paragraph{Synthetic language generation} We used an expert-annotated corpus, to avoid potential confounds between the typological parameters we manipulated and possible parse errors in an automatically parsed corpus. As our starting point, we took the English Penn Treebank \cite{penn}, converted to the Universal Dependencies scheme (\citealt{universal-dep}) using the Stanford converter \cite{stanford-converter}. We then manipulated the tree representations of the sentences in the corpus to generate parametrically modified English corpora, varying in case systems, agreement patterns, and order of core elements. For each parametric version of English, we recorded the verb-argument relations within each sentence, and created a labeled dataset.  We exposed our models to sentences from which one of the verbs was omitted, and trained them to predict the plurality of the arguments of the unseen verb. The following paragraph describes the process of collecting verb-argument relations; a detailed discussion of the parametric generation process for agreement marking, word order and case marking is given in the corresponding sections. We have made our synthetic language generation code publicly available.\footnote{\href{https://github.com/Shaul1321/rnn_typology}{https://github.com/Shaul1321/rnn\_typology} }

\paragraph{Argument collection} We created a labeled agreement prediction dataset by first collecting verb-arguments relations from the parsed corpus. We collected nouns, proper nouns, pronouns, adjectives, cardinal numbers and relative pronouns connected to a verb (identified by its part-of-speech tag) with an \textit{nsubj}, \textit{nsubjpass} or \textit{dobj} dependency edge, and record the plurality of those arguments. Verbs that were the head of a clausal complement without a subject (\textit{xcomp} dependencies) were excluded. We recorded the plurality of the dependents of the verb regardless of whether the tense and person of the verb condition agreement in English (that is, not only in third-person present-tense verbs). For relative pronouns that function as subjects or objects, we recorded the plurality of their referent; for instance, in the phrase \textit{Treasury bonds, which pay lower interest rates}, we considered the verb \textit{pay} to have a plural subject. 

\begin{table}[t]
\setcounter{table}{0}

\resizebox{\linewidth}{!}{
\begin{tabular}{llll}
\toprule
Prediction & Subject & Object & Object \\
                          task & accuracy  & accuracy     & recall      \\ \midrule
Subject                    & $94.7\pm0.3$  & -            & -           \\
Object                     & -         & $88.9\pm0.26$    & $81.8\pm1.4$    \\
Joint                      & $95.7\pm0.23$ & $90.0\pm0.1$     & $85.4\pm2.3$    \\ \bottomrule
\end{tabular}
}

\caption{\label{tbl:polypersonal-full-marking} Results of the polypersonal agreement experiments. ``Joint'' refers to multitask prediction of subject and object plurality.}

\vspace{\baselineskip}

\centering
\resizebox{0.7\linewidth}{!}{
\begin{tabular}{lll}
\toprule
& Singular & Plural \\
\midrule
Subject & -kar & -kon \\
Object & -kin & -ker \\
Indirect Object & -ken & -kre \\
\bottomrule
\end{tabular}
}


\caption{\label{tbl:suffixes} Case suffixes used in the experiments. Verbs are marked by a concatenation of the suffixes of their corresponding arguments.}

\end{table}

\paragraph{Prediction task} We experimented both with prediction of one of the arguments of the verb (subject or object), and with a joint setting in which the model predicted both arguments of each verb. Consider, for example, the prediction problem \ref{ex:seen_sent_polypersonal} (the verb in the original sentence was \textit{gave}):

\ex.\label{ex:seen_sent_polypersonal}The state $\langle \textnormal{verb} \rangle$  CenTrust 30 days to sell the Rubens .

In the joint prediction setting the system is expected to make the prediction $\langle \textnormal{subject: singular, object: plural} \rangle$. For each argument, the model predicts one of three categories: \textsc{singular}, \textsc{plural} or \textsc{none}. The \textsc{none} label was used in the object prediction task for intransitive verbs, which do not have an object; it was never used in the subject prediction task.

\paragraph{Model} We used bidirectional LSTMs with 150 hidden units. The bidirectional LSTM's representation of the left and right contexts of the verb was fed into a multilayer perceptron (MLP) with two hidden layers of sizes 100 and 50. We used independent MLPs to predict subject and object plurality. To capture morphological information, words were represented as the sum of the word embedding and embeddings of the character $n$-grams that made up the word.\footnote{Specifically, let $E_t$ and $E_{ng}$ be word and $n$-gram embedding matrices, and let $t_{w}$ and $NG_w$ be the word and the set of all $n$-grams of lengths 1 to 5, for a given word $w$. The final vector representation of $w$, $e_w$, is given by $e_w$ = $E_{t}[t]$ + $\sum\nolimits_{\textit{ng} \in \textit{NG}_w} E_{\textit{ng}}[\textit{ng}]$.}

The model (including the embedding layer) was trained end-to-end  using the Adam optimizer \cite{adam}. For each of the experiments described in the paper, we trained four models with different random initializations; we report averaged results alongside standard deviations.

\section{Polypersonal agreement \label{polypersonal-section}}

In languages with polypersonal agreement, verbs agree not only with their subject (as in English), but also with their direct object. Consider the following Basque example:\footnote{The verb in Basque agrees with the indirect object as well. In preliminary experiments, the recall of models trained on indirect object prediction was very low, due to the small number of indirect objects in the training corpus; we therefore do not include this task.}
\ex. \label{ex:eman}
\gll        \emph{Kutxazain-ek}  \emph{bezeroa-ri} \emph{liburu-ak} \emph{eman dizkiote}\\
 cashier-{\sc pl.erg} customer-{\sc sg.dat} book-{\sc pl.abs} {gave they-them-to-her/him} \\
        \trans The cashiers gave the books to the customer.

\noindent Information about the grammatical role of certain constituents in the sentence may disambiguate the function of others; most trivially, if a word is the subject of a given verb, it cannot simultaneously be its object. The goal of the present experiment is to determine whether jointly predicting both object and subject plurality improves the overall performance of the model. 

\paragraph{Corpus creation} In sentences with multiple verbs, agreement markers on verbs other than the prediction target could plausibly help predict the features on the target verb. In a preliminary experiment, we did not observe clear differences between different verb marking schemes (e.g., avoiding marking agreement on verbs other than the prediction target). We thus opted for full marking in all experiments: verbs are modified with suffixes that encode the number of all their arguments (see Figure~\ref{tbl:output-sample}). The suffixes we used for verbs are a concatenation of the respective case suffixes of their arguments (Table~\ref{tbl:suffixes}). For consistency, we remove  plurality markers from English verbs before adding our suffixes (for example, by replacing \textit{has} with \textit{have}).

\paragraph{Single task results}

The basic results are summarized in Table~\ref{tbl:polypersonal-full-marking}. Recall is calculated as the proportion of the sentences with a direct object for which the model predicted either \textsc{singular} or \textsc{plural}, but not \textsc{none}. Since all verbs included in the experiment had a subject, subject recall was 100\% and is therefore not reported. 

Plurality prediction accuracy was higher for subjects than objects. Recall for object prediction was 81.8\%, indicating that in many sentences the model was unable to identify the direct object. The lower performance on object plurality prediction is likely to be due to the fact that only about third of the sentences contain a direct object. This hypothesis is supported by the results of a preliminary experiment, in which the model was trained only on transitive sentences (with a direct object). Transitive-only training led to a reversal of the pattern: object plurality was predicted with higher accuracy than subject plurality. We conjecture that this is due to the fact that most noun modifiers in English follow the head, making the head of the object, which in general determines the plurality of the phrase, closer on average to the verb than the head of the subject (see Table~\ref{tbl:word-orders-results} below).

The accuracy we report for subject prediction, 94.7\%, is lower than the accuracy of over 99\%  reported by \citet{linzen2016assessing}. This may be due to one of several reasons. First, our training set was smaller: $\sim$35,000 sentences in our treebank corpus compared to $\sim$121,000 in their automatically parsed corpus. Second, sentences in the Wall Street Journal corpus may be more syntactically complex on average than sentences in Wikipedia, making it more challenging to identify the verb's arguments. Finally, we predicted agreement in all tenses, whereas \citet{linzen2016assessing} limited their study to the present tense (where English does in fact show agreement); it may be the case that sentences with past tense verbs are on average more complex than those with present tense verbs, regardless of the corpus.

\paragraph{Multitask training} Accuracy was higher in the joint setting: polypersonal agreement prediction is easier for the model. Subject prediction accuracy rose from 94.7\% to 95.7\%, object precision was slightly higher (90.0\% compared to 88.9\%), and object recall was significantly higher, increasing from 81.8\% to 85.4\%. We hypothesize that supervision signals from the prediction of both arguments lead to more robust abstract syntactic representations that transfer across the two tasks \cite{enguehard2017exploring}; for example, the model may be better able to identify the head of a noun phrase, regardless of whether it is the subject or the object. These findings suggest that when training on an auxiliary agreement prediction task in order to improve a language model's syntactic performance, additional supervision---in the form of predicting both subject and object---may be beneficial.

\section{Order of core elements \label{order-section}}

Languages vary in the typical order of the core elements of a clause: the subject, the object and the verb \cite{dryer2013order}. For example, whereas in English the canonical order is Subject-Verb-Object (SVO, \textit{The priests are reading the book}), in Irish it is Verb-Subject-Object (VSO, \citealt{dillon1961teach}):

\ex. 
\gll        Léann [na sagairt] [na leabhair]. \\
 read.\textsc{pres} the.\textsc{pl} priest.\textsc{pl} the.\textsc{pl} book.\textsc{pl} \\
        \trans `The priests are reading the books.'

While there are six possible orderings of these three elements, in most human languages the subject precedes both the object and the verb: about 86.5\% of the languages use either SOV or SVO orders, 9\% of the languages use VOS order, and OVS and OSV languages are extremely rare \cite{word-order-statistics}. 

To test whether RNNs have inductive biases favoring certain word orders over others, we created synthetic versions of English with all six possible orders of core elements. While natural languages often allow at least a limited degree of word order flexibility, our experiments used a simplified setting in which word order was either completely fixed (e.g., always SVO) or fully flexible, where one of the six orders was selected uniformly at random for each sentence in the corpus (the same order is used for all of the clauses of the sentence).

\subsection{Corpus creation}

Given a dependency parse for a sentence, we modulated the order of the subject and object nodes with respect to their verb. When changing the position of an argument node, we moved the entire subtree rooted in that node, including verbs and other arguments in this subtree. In the permutation process, we moved to the subject position not only nominal subjects (\textit{nsubj} and \textit{nsubjpass} edges in UD), but also clausal subjects (\textit{csubj} edges). Similarly, we moved to the object position not only nominal objects (\textit{dobj} edge), but also clausal complements (\textit{ccomp} and \textit{xcomp}).

We kept negations, adverbial modifiers, particles and auxiliaries in their original position with respect to the verb. Other non-core dependents of the verb (i.e. not the subject or the object), such as prepositional phrases, were placed according to their original position relative to the verb. For instance, in the clause \textit{the broker took them out for lunch}, the phrase \textit{for lunch} appeared directly following the verb and the arguments of the subtree in which it resides (\textit{took}, \textit{them}, \textit{the broker}) in all word orders, reflecting its original position relative to the verb \textit{took} (see Figure~\ref{tbl:output-sample}). Relative pronouns and complementizers remained in their original position.\footnote{For example, the result of transforming \ref{ex:that_orig} to VSO word order was \ref{ex:that_correct} rather than \ref{ex:that_wrong}:

\vspace{0.2cm}   
\begin{minipage}{\linewidth}
\ex.But these are not the differences that make headlines.\label{ex:that_orig}
    
        \ex.But are these not the differences that make headlines.\label{ex:that_correct}
       
        \ex. But are these not the differences make headlines that.\label{ex:that_wrong}
        
\end{minipage}
}

\begin{table*}[t]

  \resizebox{0.95\textwidth}{!}{\begin{minipage}{\textwidth}

\begin{center}

\resizebox{0.8\linewidth}{!}{
\begin{tabular}{lrlrll}
\toprule
 & \multicolumn{2}{c}{Subject} & \multicolumn{3}{c}{Object} \\
 \cmidrule(lr){2-3} \cmidrule(lr){4-6}
Order     & \% Attractors & Accuracy & \% Attractors & Accuracy & Recall      \\ \midrule
Unchanged & $11.56$                             & $95.7\pm0.23$         & $2.55$                            & $90.0\pm0.1$ & $85.4\pm2.37$  \\ 
SVO       & $13.16$                             & $95.4\pm0.41$         & $2.6$                            & $87.3\pm0.23$ & $80.0\pm2.61$ \\
SOV       & $78.12$                            & $90.6\pm0.37$         & $17.04$                            & $79.2\pm0.78$ & $63.3\pm4.62$ \\
VOS       & $69.50$                             & $89.5\pm0.54$         & $2.57$                            & $84.0\pm0.39$ & $77.8\pm3.68$ \\
VSO       & $6.65$                             & $95.1\pm0.12$         & $16.09$                            & $82.8\pm0.7$ & $70.0\pm1.91$  \\ 
OSV       & $14.81$                             & $93.6\pm0.23$         & $30.00$                           & $78.9\pm0.17$ & $63.5\pm4.59$ \\
OVS       & $8.13$                             & $95.7\pm0.37$         & $16.42$                            & $83.7\pm0.32$ & $72.8\pm1.58$ \\
Flexible    & $32.24$                             & $88.6\pm0.43$         & $14.44$                            & $74.1\pm0.70$ & $60.2\pm3.24$ \\\bottomrule
\end{tabular}
}
\caption{\label{tbl:word-orders-results} Subject and object plurality prediction for different word orders (recall for the subject is 100\% and is not indicated). The \% attractors columns indicate the percentage of sentences containing verb-argument attractors. The number are averaged over four runs and the error interval represents the standard deviation.}

\end{center}
      \end{minipage}}

\end{table*}

In all experiments in this section, we trained the model to jointly predict the plurality of the subject and the object. For consistency across the object and subject plurality prediction tasks, we used the polypersonal agreement markers on all verbs in the sentence (except, of course, for the prediction target, which was withheld completely). For example, in the OVS version of the sentence presented in Figure~\ref{tbl:output-sample}, the input was \ref{ex:seen_sent_reordered}, where \textit{kon} marks the fact that \textit{say} has a plural subject:

\ex. 
		 them 	$\langle \textnormal{verb} \rangle$ out frequently the broker for lunch say\textit{kon} they .
         \label{ex:seen_sent_reordered}

\subsection{Results}
\label{sub:wordorderresults}

Performance varied significantly across word orders (Table~\ref{tbl:word-orders-results}). Subject plurality prediction accuracy was inversely correlated with the frequency of attractors (intervening nouns of the opposite plurality) in the language: accuracy was lowest for subject prediction in the VOS and SOV languages, in which objects intervene between the subject and the verb (Figure~\ref{fig:attractors}). The degraded performance in these languages is consistent with the attraction effects found in previous studies of agreement in natural languages \cite{linzen2016assessing,gulordava2018LMagreement}, and support the hypothesis that RNNs have an inductive bias favoring dependencies with recent elements; we test this hypothesis in a more controlled way in \S\ref{sec:noobjects}.

\begin{figure}[h]
\centering
\setcounter{figure}{1}
\includegraphics[scale=0.2]{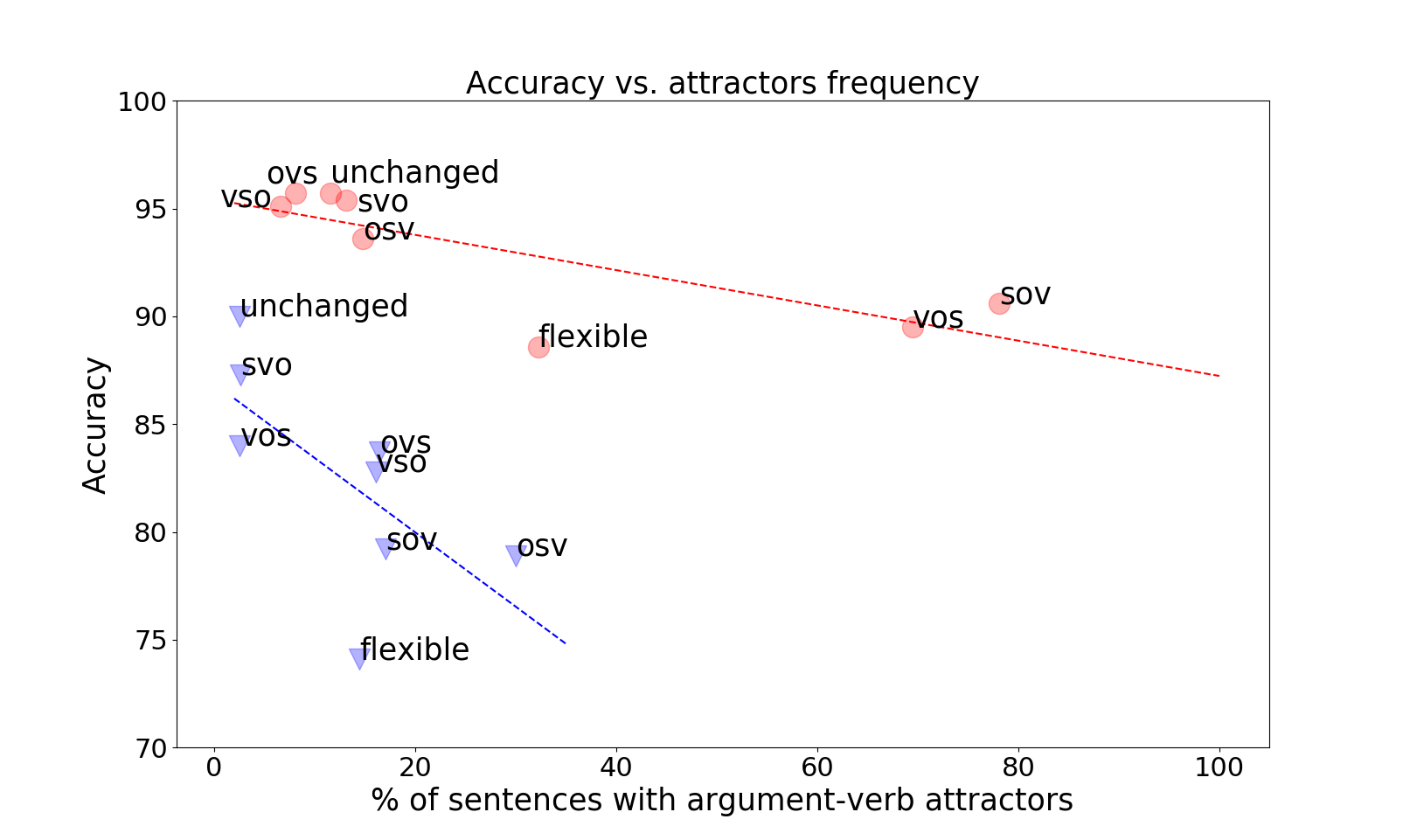}
\caption{\label{fig:attractors} Subject and object plurality prediction accuracy as a function of the percentage of sentences with attractors that are arguments of the verb. Red circles represent subject prediction and blue triangles represent object prediction. $R^{2}$: 0.61 for subject, 0.43 for object.}
\end{figure}

Attractors affected object prediction accuracy as well. The highest accuracy among the synthetic languages was in the SVO language and the worst performance observed in the OSV language. As in \S\ref{polypersonal-section}, subjects were easier to predict than objects, likely because all verbs in the training set had a subject, but only 35\% had an object. 

Flexible word order was especially challenging for the model, with a subject plurality prediction accuracy of 88.6\%, object plurality prediction accuracy of 74.1\%, and object recall of 60.2\%. This does not necessarily bear on the RNNs' inductive biases: flexible word order without case marking would make it difficult for any learner to infer syntactic relations. Without overt cues, the model must resort to selectional restrictions (e.g., in \textit{the apples ate the man}, the only plausible subject is \textit{the man}), but those are difficult to learn from a small corpus. What's more, some sentences are truly ambiguous when there are no case markers or word order cues; this happens for example when both arguments are animate, as in \textit{the lawyer saw the doctor} \cite{gibson2013noisy,ettinger2018assessing}. 

\subsection{Withholding direct objects in training}
\label{sec:noobjects}

The previous experiments suggested that the RNN has a tendency to identify the more recent argument as the subject, leading to attraction effects caused by the object. We conjectured that this is due to the fact that many verbs are intransitive, that is, have a subject but not an object. The clauses in which those verbs appear provide ambiguous evidence: they are equally compatible with a generalization in which the subject is the \textit{most recent} core element before the verb, and with a generalization in which the subject is the \textit{first} core constituent of the clause. Attraction effects suggest that the inductive bias of the RNN leads it to adopt the incorrect recency-based generalization. To test this hypothesis in a controlled way, we adopt the ``poverty of the stimulus'' paradigm \cite{wilson2006learning,culbertson2014language,mccoy2018revisiting}: we withhold all evidence that disambiguates these two hypotheses (namely, all transitive sentences), and test how the RNN generalizes to the withheld sentence type.

We used the SOV and VOS corpora described before; in both of these languages, the object intervened between the subject and the verb, potentially causing agreement attraction. Crucially, we train only on sentences \textit{without} a direct object, and test on the following three types of sentences: 

\begin{enumerate}
\item Sentences with an object of the opposite plurality from the subject (object attractor).
\item Sentences with an object of the same plurality as the subject (non-attractor object).\footnote{When the object is a noun-noun compound, it is considered a non-attractor if its head is not of the opposite plurality of the subject, regardless of the plurality of other elements. This can only make the task harder compared with the alternative of considering compound objects such as ``screen displays'' as attractors for plural subjects.}
\item Sentences without an object, but with one or more nouns of the opposite plurality intervening between the subject and the verb (non-object attractor); e.g., \textit{The gap between winners and losers will grow} is intransitive, but the plural words \textit{winners} and \textit{losers}, which are a part of a modifier of the subject, may serve as attractors for the singular subject \textit{gap}.
\end{enumerate}

\noindent The results are shown in Table~\ref{tbl:subject-generalization}. Withholding direct objects during training dramatically degraded the performance of the model on sentences with an object attractor: the accuracy decreased from 90.6\% for the model trained on the full SOV corpus (Table~\ref{tbl:word-orders-results}) to 60.0\% for the model trained only on intransitive sentences from the same corpus. There was an analogous drop in performance in the case of VOS (89.5\% compared to 48.3\%). By contrast, attractors that were not core arguments, or objects that were not attractors, did not hurt performance in a comparable way. This suggests that in our poverty of the stimulus experiments RNNs were able to distinguish between core and non-core elements, but struggled on instances in which where the object directly preceded the verb (the instances that were withheld in training). This constitutes strong evidence for the RNN's recency bias: our models extracted the generalization that subjects directly precede the verb, even though the data were equally compatible with the generalization that the subject is the first core argument in the clause.

These findings align with the results of \citet{UrvashiContextLength}, who demonstrated that RNN language models are more sensitive to perturbations in recent input words compared with perturbations to more distant parts of the input. While in their case the model's recency preference can be a learned property (since recent information is more relevant for the task of next-word prediction), our experiment focuses on the inherent inductive biases of the model, as the cues that are necessary for differentiating between the two generalizations were absent in training. 

\subsection{Discussion}

Our reordering manipulation was limited to core element (subjects, objects and verbs). Languages also differ in word order inside other types of phrases, including noun phrases (e.g., does an adjective precede or follow the noun?),  adpositional phrases (does the language use prepositions or postpositions?), and so on. \citet{greenberg1963} pointed out correlations between head-modifier orders across phrase categories; while a significant number of exceptions exist, these correlations have motivated proposals for a language-wide setting of a Head Directionality Parameter \cite{stowell1981origins,baker2001atoms}. In future work, we would like to explore whether consistent reordering across categories improves the model's performance.

\begin{table}[t]

  \begin{minipage}{\linewidth}

\resizebox{\linewidth}{!}{
\begin{tabular}{cccc}
\toprule        
& Object & Object & Non-object \\
& (attractor) & (non attractor) & attractor \\
\midrule
SOV & $60.3\pm 3.7$ & $92.8 \pm 0.3$              & $79.2 \pm 3$               \\
VOS & $48.3\pm 2.3$ & $94.0\pm 2.3$              & $83.1\pm1.1$             \\ \bottomrule
\end{tabular}
}
\caption{\label{tbl:subject-generalization} Subject prediction accuracy in the ``poverty of the stimulus'' paradigm of Section~\ref{sec:noobjects}, where transitive sentences were withheld during training. Numbers are averaged over four runs and the error interval represents the standard deviation.}
      \end{minipage}

\end{table}

In practice, even languages with a relatively rigid word order almost never enforce this order in every clause. The order of elements in English, for example, is predominately SVO, but constructions in which the verb precedes the subject do exist, e.g., \textit{Outside were three police officers}. Other languages are considerably more flexible than English \cite{dryer2013order}. Given that word order flexibility makes the task more difficult, our setting is arguably simpler than the task the model would face when learning a natural language.

The fact that the agreement dependency between the subject and the verb was more challenging to establish in the SOV order compared to the SVO order is consistent with the hypothesis that SVO languages make it easier to distinguish the subject from the object \cite{gibson2013noisy}; indeed, to compensate for this issue, SOV languages more frequently employ case marking (Matthew Dryer, quoted in \citealt{gibson2013noisy}). 

There was not a clear relationship between the prevalence of a particular word order in the languages of the world and the difficulty that our models experienced with that order. The model performed best on the OVS word order, which is present in a very small number of languages ($\sim$1\%). SOV languages were more difficult for our RNNs to learn than SVO languages, even though SOV languages are somewhat \textit{more} common \cite{dryer2013order}. These results weakly support functional explanations of these typological tendencies; such explanations appeal to communicative efficiency considerations rather than learning biases \cite{maurits2010wordorder}. Of course, since the inductive biases of humans and RNNs are likely to be different in many respects, our results do not rule out the possibility that the distribution of word orders is driven by a human learning bias after all.

\section{Overt morphological case systems \label{cases-section}}

\begin{table*}[t]

  \resizebox{0.99\textwidth}{!}{\begin{minipage}{\textwidth}

\begin{tabular}{lllllll}
\toprule
\multirow{2}{*}{Case system}
                                      & \multicolumn{2}{c}{Flexible word order}                                            & \multicolumn{2}{c}{VOS}                                                 & \multicolumn{2}{c}{OVS}                                                 \\ \cmidrule(lr){2-3} \cmidrule(lr){4-5}
                                      \cmidrule(lr){6-7}
                                      & \multicolumn{1}{c}{Subject A} & \multicolumn{1}{c}{Object A/R} & \multicolumn{1}{c}{Subject A} & \multicolumn{1}{c}{Object A/R} & \multicolumn{1}{c}{Subject A} & \multicolumn{1}{c}{Object A/R} \\ \midrule
Unambiguous                  & $99.2\pm0.5$                                   & $98.7\pm0.2$                              & $98.9\pm0.2$                                   & $99.5\pm0.1$                               & $99.5\pm0.2$                                   & $98.6\pm0.3$                              \\ 

   & & /$98.0\pm0.5$  & &  /$99.1\pm0.1$ & &  /$98.4\pm0.6$  \\ 

Syncretic                 & $99.3\pm0.2$                                   & $93.6\pm0.4$                              & $99.1\pm 0.2$                                   & $97.1\pm0.2$                               & $99.4\pm0.1$                                   & $97.8\pm0.2$                               \\

   & & /$88.9\pm1.7$  & &  /$95.0\pm1.1$ & &  /$97.4\pm1.2$ \\
Argument marking                  & $96.0\pm0.3$                                   & $86.1\pm0.9$                              & $96.9\pm 0.1$                                   & $93.6\pm0.1$                               &  $99.6\pm0.1$                                   & $96.8\pm0.1$                               \\ 
   & & /$79.7\pm4.9$  & &  /$89.8\pm2.4$ & & /$95.5\pm0.5$ \\

\bottomrule
\end{tabular}

\caption{\label{tbl:cases-results} Accuracy (A) and recall (R) in predicting subject and object agreement with different case systems.}
      \end{minipage}}

\end{table*}

The vast majority of noun phrases in English are not overtly marked for grammatical function (case), with the exception of pronouns; e.g., the first-person singular pronoun is \textit{I} when it is a subject and \textit{me} when it is an object. Other languages mark case on most nouns. Consider, for example, the following example from Russian:\footnote{The standard grammatical term for these cases are nominative (for subject) and accusative (for object); we use \textsc{subject} and \textsc{object} for clarity.}

\ex.\a.\gll ya kupil knig-u.\\
    I bought book-\textsc{object}\\
    \trans `I bought the book.'
    \b.\gll knig-a ischezla.\\
    book-\textsc{subject} disappeared\\
    \trans `The book disappeared.'

\noindent Overt case marking reduces ambiguity and facilitates parsing languages with flexible word order. To investigate the influence of case on agreement prediction---and on the ability to infer sentence structure---we experimented with different case systems. In all settings, we used ``fused'' suffixes, which encode both plurality and grammatical function. We considered three case systems (see Figure~\ref{tbl:output-sample}):
\begin{enumerate}
    \item An unambiguous case system, with a unique suffix for each combination of number and grammatical function.
    \item A partially syncretic (ambiguous) case system, in which the same suffix was attached to both singular subjects and plural objects (modeled after Basque). 
    \item A fully syncretic case system (argument marking only): the suffix indicated only the plurality of the argument, regardless of its grammatical function (cf. subject/object syncretism in Russian neuter nouns).
\end{enumerate}
In the typological survey reported in \citet{baermann2013syncretism}, 62\% of the languages had no or minimal case marking, 20\% had syncretic case systems, and 18\% had case systems with no syncretism.

\paragraph{Corpus creation}

The suffixes we used are listed in Table~\ref{tbl:suffixes}. We only attached the suffix to the head of the relevant argument; adjectives and other modifiers did not carry case suffixes. The same suffix was used to mark plurality/case on noun and the agreement features on the verb; e.g., if the verb \textit{eat} had a singular subject and plural object, it appeared as eat\textit{karker} (the singular subject suffix was \textit{kar} and the plural object suffix was \textit{ker}). We stripped off plurality and case markers from the original English noun phrases before adding these suffixes. 

\paragraph{Setup} We evaluated the interaction between different case marking schemes and three word orders: flexible word order and the two orders on which the model achieved the best (OVS) and worst (VOS) subject prediction accuracy. We train one model for each combination of case system and word order. We jointly predicted the plurality of subject and the object.

\paragraph{Results and analysis}

The results are summarized in Table~\ref{tbl:cases-results}. Unambiguous case marking dramatically improved subject and object plurality prediction compared with the previous experiments; accuracy was above 98\% for all three word orders. Partial syncretism hurt performance somewhat relative to the unambiguous setting (except with flexible word order), especially for object prediction. The fully syncretic case system, which marked only the plurality of the head of each argument, further decreased performance. At the same time, even this limited marking scheme was helpful: accuracy in the most challenging setting, flexible word order (subject: 96.0\%; object: 86.1\%), was not very different from the results on unmodified English (95.7\% and 90.0\%). This contrasts with the poor results on the flexible setting without cases (subject: 88.6\%; object: 60.2\%). On the rigid orders, a fully syncretic system still significantly improved agreement prediction. The moderate effect of case syncretism on performance suggests that most of the benefits of case marking stems from the overt marking of the heads of all arguments.

Overall, these results are consistent with the observation that languages with explicit case marking tend to allow a more flexible word orders compared with languages such as English that make use of word order to express grammatical function of words. 

\section{Related work}
Our approach of constructing synthetic languages by parametrically modifying parsed corpora for natural languages is closely inspired by  \citet{wang2016galactic} (see also \citealt{wang2017fine}). While they trained a model to mimic the POS tags order-statistics of the target language, we manually modified the parsed corpora; this allows us to control for selected parameters, at the expense of reducing generality.

Simpler synthetic languages (not based on natural corpora) have been used in a number of recent studies to examine the inductive biases of different neural architectures \cite{bowman2015tree,lake2018generalization,mccoy2018revisiting}. In another recent study, \citet{cotterell2018languagemodel} measured the ability of RNN and $n$-gram models to perform character-level language modeling in a sample of languages, using a parallel corpus; the main typological property of interest in that study was morphological complexity. Finally, a large number of studies, some mentioned in the introduction, have used syntactic prediction tasks to examine the generalizations acquired by neural models (see also \citealt{bernardy2017using,futrell2018rnns,lau2017grammaticality,conneau2018cram,ettinger2018assessing,jumelet2018language}). 

\section{Conclusions}

We have proposed a methodology for generating parametric variations of existing languages and evaluating the performance of RNNs in syntactic feature prediction in the resulting languages. We used this methodology to study the grammatical inductive biases of RNNs, assessed whether certain grammatical phenomena are more challenging for RNNs to learn than others, and began to compare these patterns with the linguistic typology literature.

In our experiments, multitask training on polypersonal agreement prediction improved performance, suggesting that the models acquired syntactic representations that generalize across argument types (subjects and objects). Performance varied significantly across word orders. This variation was not correlated with the frequency of the word orders in the languages of the world. Instead, it was inversely correlated with the frequency of attractors, demonstrating a recency bias. Further supporting this bias, in a poverty-of -the-stimulus paradigm, where the data were equally consistent with two generalizations---first, the generalization that the subject is the first argument in the clause, and second, the generalization that the subject is the most recent argument preceding the verb---RNNs adopted the recency-based generalization. Finally, we found that overt case marking on the heads of arguments dramatically improved plurality prediction performance, even when the case system was highly syncretic.

Agreement feature prediction in some of our synthetic languages is likely to be difficult not only for RNNs but for many other classes of learners, including humans. For example, agreement in a language with very flexible word order and without case marking is impossible to predict in many cases (see \S\ref{sub:wordorderresults}), and indeed such languages are very rare. In future work, a human experiment based on the agreement prediction task can help determine whether the difficulty of our languages is consistent across humans and RNNs.

\section*{Acknowledgement}
This work is supported by the Israeli Science Foundation (grant number 1555/15) and by Theo Hoffenberg, the founder \& CEO of Reverso.

\bibliographystyle{acl_natbib}
\bibliography{rnn_typology}

\begin{thebibliography}{39}
\expandafter\ifx\csname natexlab\endcsname\relax\def\natexlab#1{#1}\fi

\bibitem[{Baerman and Brown(2013)}]{baermann2013syncretism}
Matthew Baerman and Dunstan Brown. 2013.
\newblock \href {https://wals.info/chapter/28} {Case syncretism}.
\newblock In Matthew~S. Dryer and Martin Haspelmath, editors, \emph{The World
  Atlas of Language Structures Online}. Max Planck Institute for Evolutionary
  Anthropology, Leipzig.

\bibitem[{Baker(2001)}]{baker2001atoms}
Mark~C. Baker. 2001.
\newblock \emph{The atoms of language: The mind's hidden rules of grammar}.
\newblock Basic Books, New York.

\bibitem[{Bernardy and Lappin(2017)}]{bernardy2017using}
Jean-Philippe Bernardy and Shalom Lappin. 2017.
\newblock Using deep neural networks to learn syntactic agreement.
\newblock \emph{LiLT (Linguistic Issues in Language Technology)}, 15.

\bibitem[{Bowman et~al.(2015)Bowman, Manning, and Potts}]{bowman2015tree}
Samuel~R. Bowman, Christopher~D. Manning, and Christopher Potts. 2015.
\newblock Tree-structured composition in neural networks without
  tree-structured architectures.
\newblock In \emph{{Proceedings of the NIPS Workshop on Cognitive Computation:
  Integrating Neural and Symbolic Approaches}}.

\bibitem[{Chomsky(1981)}]{chomsky1981lectures}
Noam Chomsky. 1981.
\newblock \emph{{L}ectures on {G}overnment and {B}inding}.
\newblock {F}oris, Dordrecht.

\bibitem[{Chowdhury and Zamparelli(2018)}]{chowdhury2018rnn}
Shammur~Absar Chowdhury and Roberto Zamparelli. 2018.
\newblock \href {http://aclweb.org/anthology/C18-1012} {{RNN} simulations of
  grammaticality judgments on long-distance dependencies}.
\newblock In \emph{{Proceedings of the 27th International Conference on
  Computational Linguistics}}, pages 133--144. Association for Computational
  Linguistics.

\bibitem[{Conneau et~al.(2018)Conneau, Kruszewski, Lample, Barrault, and
  Baroni}]{conneau2018cram}
Alexis Conneau, Germ{\'a}n Kruszewski, Guillaume Lample, Lo{\"\i}c Barrault,
  and Marco Baroni. 2018.
\newblock \href {http://aclweb.org/anthology/P18-1198} {What you can cram into
  a single vector: Probing sentence embeddings for linguistic properties}.
\newblock In \emph{{Proceedings of the 56th Annual Meeting of the Association
  for Computational Linguistics (Volume 1: Long Papers)}}, pages 2126--2136.
  Association for Computational Linguistics.

\bibitem[{Cotterell et~al.(2018)Cotterell, Mielke, Eisner, and
  Roark}]{cotterell2018languagemodel}
Ryan Cotterell, Sebastian~J. Mielke, Jason Eisner, and Brian Roark. 2018.
\newblock \href {https://doi.org/10.18653/v1/N18-2085} {Are all languages
  equally hard to language-model?}
\newblock In \emph{{Proceedings of the 2018 Conference of the North American
  Chapter of the Association for Computational Linguistics: Human Language
  Technologies, Volume 2 (Short Papers) }}, pages 536--541. Association for
  Computational Linguistics.

\bibitem[{Culbertson and Adger(2014)}]{culbertson2014language}
Jennifer Culbertson and David Adger. 2014.
\newblock Language learners privilege structured meaning over surface
  frequency.
\newblock \emph{Proceedings of the National Academy of Sciences}, page
  201320525.

\bibitem[{Dillon and Ó~Cróinin(1961)}]{dillon1961teach}
Myles Dillon and Donncha Ó~Cróinin. 1961.
\newblock \emph{Teach Yourself Irish}.
\newblock The English Universities Press Ltd., London.

\bibitem[{Dryer(2013)}]{dryer2013order}
Matthew~S. Dryer. 2013.
\newblock \href {https://wals.info/chapter/81} {Order of subject, object and
  verb}.
\newblock In Matthew~S. Dryer and Martin Haspelmath, editors, \emph{The World
  Atlas of Language Structures Online}. Max Planck Institute for Evolutionary
  Anthropology, Leipzig.

\bibitem[{Enguehard et~al.(2017)Enguehard, Goldberg, and
  Linzen}]{enguehard2017exploring}
\'{E}mile Enguehard, Yoav Goldberg, and Tal Linzen. 2017.
\newblock \href {http://aclweb.org/anthology/K17-1003} {Exploring the syntactic
  abilities of {RNNs} with multi-task learning}.
\newblock In \emph{{Proceedings of the 21st Conference on Computational Natural
  Language Learning (CoNLL 2017)}}, pages 3--14.

\bibitem[{Ettinger et~al.(2018)Ettinger, Elgohary, Phillips, and
  Resnik}]{ettinger2018assessing}
Allyson Ettinger, Ahmed Elgohary, Colin Phillips, and Philip Resnik. 2018.
\newblock \href {http://aclweb.org/anthology/C18-1152} {Assessing composition
  in sentence vector representations}.
\newblock In \emph{{Proceedings of the 27th International Conference on
  Computational Linguistics}}, pages 1790--1801. Association for Computational
  Linguistics.

\bibitem[{Futrell et~al.(2018)Futrell, Wilcox, Morita, and
  Levy}]{futrell2018rnns}
Richard Futrell, Ethan Wilcox, Takashi Morita, and Roger Levy. 2018.
\newblock {RNNs} as psycholinguistic subjects: Syntactic state and grammatical
  dependency.
\newblock \emph{arXiv preprint arXiv:1809.01329}.

\bibitem[{Gibson et~al.(2013)Gibson, Piantadosi, Brink, Bergen, Lim, and
  Saxe}]{gibson2013noisy}
Edward Gibson, Steven~T. Piantadosi, Kimberly Brink, Leon Bergen, Eunice Lim,
  and Rebecca Saxe. 2013.
\newblock A noisy-channel account of crosslinguistic word-order variation.
\newblock \emph{Psychological Science}, 24(7):1079--1088.

\bibitem[{Giulianelli et~al.(2018)Giulianelli, Harding, Mohnert, Hupkes, and
  Zuidema}]{giulianelli2018hood}
Mario Giulianelli, Jack Harding, Florian Mohnert, Dieuwke Hupkes, and Willem
  Zuidema. 2018.
\newblock \href {http://aclweb.org/anthology/W18-5426} {Under the hood: Using
  diagnostic classifiers to investigate and improve how language models track
  agreement information}.
\newblock In \emph{{Proceedings of the 2018 EMNLP Workshop BlackboxNLP:
  Analyzing and Interpreting Neural Networks for NLP}}, pages 240--248.
  Association for Computational Linguistics.

\bibitem[{Greenberg(1963)}]{greenberg1963}
Joseph~H. Greenberg. 1963.
\newblock Some universals of grammar with particular reference to the order of
  meaningful elements.
\newblock In Joseph~H. Greenberg, editor, \emph{Universals of language}, pages
  73--113. MIT Press, Cambridge, MA.

\bibitem[{Gulordava et~al.(2018)Gulordava, Bojanowski, Grave, Linzen, and
  Baroni}]{gulordava2018LMagreement}
Kristina Gulordava, Piotr Bojanowski, Edouard Grave, Tal Linzen, and Marco
  Baroni. 2018.
\newblock \href {https://aclanthology.info/papers/N18-1108/n18-1108} {Colorless
  green recurrent networks dream hierarchically}.
\newblock In \emph{Proceedings of the 2018 Conference of the North American
  Chapter of the Association for Computational Linguistics: Human Language
  Technologies, {NAACL-HLT} 2018, New Orleans, Louisiana, USA, June 1-6, 2018,
  Volume 1 (Long Papers)}, pages 1195--1205.

\bibitem[{Jumelet and Hupkes(2018)}]{jumelet2018language}
Jaap Jumelet and Dieuwke Hupkes. 2018.
\newblock \href {http://aclweb.org/anthology/W18-5424} {Do language models
  understand anything? on the ability of lstms to understand negative polarity
  items}.
\newblock In \emph{Proceedings of the 2018 EMNLP Workshop BlackboxNLP:
  Analyzing and Interpreting Neural Networks for NLP}, pages 222--231.
  Association for Computational Linguistics.

\bibitem[{Khandelwal et~al.(2018)Khandelwal, He, Qi, and
  Jurafsky}]{UrvashiContextLength}
Urvashi Khandelwal, He~He, Peng Qi, and Dan Jurafsky. 2018.
\newblock \href {https://aclanthology.info/papers/P18-1027/p18-1027} {Sharp
  nearby, fuzzy far away: How neural language models use context}.
\newblock In \emph{Proceedings of the 56th Annual Meeting of the Association
  for Computational Linguistics, {ACL} 2018, Melbourne, Australia, July 15-20,
  2018, Volume 1: Long Papers}, pages 284--294.

\bibitem[{Kingma and Ba(2014)}]{adam}
Diederik~P. Kingma and Jimmy Ba. 2014.
\newblock \href {http://arxiv.org/abs/1412.6980} {Adam: {A} method for
  stochastic optimization}.
\newblock \emph{arXiv preprint 1412.6980}.

\bibitem[{Kuncoro et~al.(2018)Kuncoro, Dyer, Hale, Yogatama, Clark, and
  Blunsom}]{kuncoro2018lstms}
Adhiguna Kuncoro, Chris Dyer, John Hale, Dani Yogatama, Stephen Clark, and Phil
  Blunsom. 2018.
\newblock \href {http://aclweb.org/anthology/P18-1132} {{LSTMs} can learn
  syntax-sensitive dependencies well, but modeling structure makes them
  better}.
\newblock In \emph{{Proceedings of the 56th Annual Meeting of the Association
  for Computational Linguistics (Volume 1: Long Papers)}}, pages 1426--1436.
  Association for Computational Linguistics.

\bibitem[{Lake and Baroni(2018)}]{lake2018generalization}
Brenden~M. Lake and Marco Baroni. 2018.
\newblock \href {http://proceedings.mlr.press/v80/lake18a.html} {Generalization
  without systematicity: On the compositional skills of sequence-to-sequence
  recurrent networks}.
\newblock In \emph{Proceedings of the 35th International Conference on Machine
  Learning, {ICML} 2018, Stockholmsm{\"{a}}ssan, Stockholm, Sweden, July 10-15,
  2018}, pages 2879--2888.

\bibitem[{Lau et~al.(2017)Lau, Clark, and Lappin}]{lau2017grammaticality}
Jey~Han Lau, Alexander Clark, and Shalom Lappin. 2017.
\newblock Grammaticality, acceptability, and probability: A probabilistic view
  of linguistic knowledge.
\newblock \emph{Cognitive Science}, 41(5):1202--1247.

\bibitem[{Linzen et~al.(2016)Linzen, Dupoux, and
  Goldberg}]{linzen2016assessing}
Tal Linzen, Emmanuel Dupoux, and Yoav Goldberg. 2016.
\newblock \href {https://transacl.org/ojs/index.php/tacl/article/view/972}
  {Assessing the ability of {LSTMs} to learn syntax-sensitive dependencies}.
\newblock \emph{Transactions of the Association for Computational Linguistics},
  4:521--535.

\bibitem[{Marcus et~al.(1993)Marcus, Santorini, and Marcinkiewicz}]{penn}
Mitchell~P. Marcus, Beatrice Santorini, and Mary~Ann Marcinkiewicz. 1993.
\newblock Building a large annotated corpus of {English}: The {Penn Treebank}.
\newblock \emph{Computational Linguistics}, 19(2):313--330.

\bibitem[{Marvin and Linzen(2018)}]{marvin2018targeted}
Rebecca Marvin and Tal Linzen. 2018.
\newblock \href {http://aclweb.org/anthology/D18-1151} {Targeted syntactic
  evaluation of language models}.
\newblock In \emph{{Proceedings of the 2018 Conference on Empirical Methods in
  Natural Language Processing}}, pages 1192--1202.

\bibitem[{Maurits et~al.(2010)Maurits, Navarro, and
  Perfors}]{maurits2010wordorder}
Luke Maurits, Danielle~J. Navarro, and Amy Perfors. 2010.
\newblock \href
  {http://papers.nips.cc/paper/4085-why-are-some-word-orders-more-common-than-others-a-uniform-information-density-account}
  {Why are some word orders more common than others? {A} uniform information
  density account}.
\newblock In \emph{Advances in Neural Information Processing Systems 23: 24th
  Annual Conference on Neural Information Processing Systems 2010. Proceedings
  of a meeting held 6-9 December 2010, Vancouver, British Columbia, Canada.},
  pages 1585--1593. Curran Associates, Inc.

\bibitem[{McCoy et~al.(2018)McCoy, Frank, and Linzen}]{mccoy2018revisiting}
R.~Thomas McCoy, Robert Frank, and Tal Linzen. 2018.
\newblock \href {http://mindmodeling.org/cogsci2018/papers/0399/0399.pdf}
  {Revisiting the poverty of the stimulus: Hierarchical generalization without
  a hierarchical bias in recurrent neural networks}.
\newblock In \emph{{Proceedings of the 40th Annual Conference of the Cognitive
  Science Society}}, pages 2093---2098.

\bibitem[{Nivre et~al.(2016)Nivre, de~Marneffe, Ginter, Goldberg, Hajic,
  Manning, McDonald, Petrov, Pyysalo, Silveira, Tsarfaty, and
  Zeman}]{universal-dep}
Joakim Nivre, Marie{-}Catherine de~Marneffe, Filip Ginter, Yoav Goldberg, Jan
  Hajic, Christopher~D. Manning, Ryan~T. McDonald, Slav Petrov, Sampo Pyysalo,
  Natalia Silveira, Reut Tsarfaty, and Daniel Zeman. 2016.
\newblock \href
  {http://www.lrec-conf.org/proceedings/lrec2016/summaries/348.html} {Universal
  dependencies v1: {A} multilingual treebank collection}.
\newblock In \emph{Proceedings of the Tenth International Conference on
  Language Resources and Evaluation {LREC} 2016, Portoro{\v{z}}, Slovenia, May
  23-28, 2016.}

\bibitem[{Ravfogel et~al.(2018)Ravfogel, Tyers, and
  Goldberg}]{ravfogel2018basque}
Shauli Ravfogel, Francis Tyers, and Yoav Goldberg. 2018.
\newblock \href {http://aclweb.org/anthology/W18-5412} {Can {LSTM} learn to
  capture agreement? the case of {Basque}}.
\newblock In \emph{{Proceedings of the 2018 EMNLP Workshop BlackboxNLP:
  Analyzing and Interpreting Neural Networks for NLP}}, pages 98--107.
  Association for Computational Linguistics.

\bibitem[{Schuster and Manning(2016)}]{stanford-converter}
Sebastian Schuster and Christopher~D. Manning. 2016.
\newblock \href
  {http://www.lrec-conf.org/proceedings/lrec2016/summaries/779.html} {Enhanced
  {English} universal dependencies: An improved representation for natural
  language understanding tasks}.
\newblock In \emph{Proceedings of the Tenth International Conference on
  Language Resources and Evaluation {LREC} 2016, Portoro{\v{z}}, Slovenia, May
  23-28, 2016.}

\bibitem[{Stowell(1981)}]{stowell1981origins}
Timothy~Angus Stowell. 1981.
\newblock \emph{Origins of phrase structure}.
\newblock Ph.D. thesis, Massachusetts Institute of Technology.

\bibitem[{Tomlin(1986)}]{word-order-statistics}
Rudolf~S. Tomlin. 1986.
\newblock \emph{Basic word order: functional principles}.
\newblock Croom Helm, London.

\bibitem[{{van Schijndel} and Linzen(2018)}]{vanschijndel2018gardenpath}
Marten {van Schijndel} and Tal Linzen. 2018.
\newblock Modeling garden path effects without explicit hierarchical syntax.
\newblock In \emph{{Proceedings of the 40th Annual Conference of the Cognitive
  Science Society}}, pages 2600--2605, Austin, TX. Cognitive Science Society.

\bibitem[{Wang and Eisner(2016)}]{wang2016galactic}
Dingquan Wang and Jason Eisner. 2016.
\newblock \href {http://aclweb.org/anthology/Q16-1035} {The galactic
  dependencies treebanks: Getting more data by synthesizing new languages}.
\newblock \emph{Transactions of the Association for Computational Linguistics},
  4:491--505.

\bibitem[{Wang and Eisner(2017)}]{wang2017fine}
Dingquan Wang and Jason Eisner. 2017.
\newblock \href {http://aclweb.org/anthology/Q17-1011} {Fine-grained prediction
  of syntactic typology: Discovering latent structure with supervised
  learning}.
\newblock \emph{Transactions of the Association for Computational Linguistics},
  5:147--161.

\bibitem[{Wilcox et~al.(2018)Wilcox, Levy, Morita, and
  Futrell}]{wilcox2018fillergap}
Ethan Wilcox, Roger Levy, Takashi Morita, and Richard Futrell. 2018.
\newblock \href {http://aclweb.org/anthology/W18-5423} {What do {RNN} language
  models learn about filler--gap dependencies?}
\newblock In \emph{{Proceedings of the 2018 EMNLP Workshop BlackboxNLP:
  Analyzing and Interpreting Neural Networks for NLP}}, pages 211--221.
  Association for Computational Linguistics.

\bibitem[{Wilson(2006)}]{wilson2006learning}
Colin Wilson. 2006.
\newblock Learning phonology with substantive bias: An experimental and
  computational study of velar palatalization.
\newblock \emph{Cognitive Science}, 30:945--982.

\end{thebibliography}
\end{document}